\documentclass[10pt,twocolumn,letterpaper]{article}

\usepackage{cvpr}
\usepackage{times}
\usepackage{epsfig}
\usepackage{graphicx}
\usepackage{amsmath}
\usepackage{amssymb}

\usepackage{bbm}
\usepackage{booktabs}
\usepackage{caption}
\usepackage{mathtools}
\usepackage{microtype}
\newcommand*\rot{\rotatebox{90}}

\usepackage[pagebackref=true,breaklinks=true,letterpaper=true,colorlinks,bookmarks=false]{hyperref}

\usepackage{colortbl}
\definecolor{Gray}{gray}{0.85}
\newcolumntype{g}{>{\columncolor{Gray}} c}

\cvprfinalcopy 

\def\cvprPaperID{3384} 

\ifcvprfinal\fi
\begin{document}

\title{Adversarial Discriminative Domain Adaptation}

\author{
  Eric Tzeng\\
  University of California, Berkeley\\
  \texttt{etzeng@eecs.berkeley.edu}\\
  \and
  Judy Hoffman\\
  Stanford University\\
  \texttt{jhoffman@cs.stanford.edu}\\
  \and
  Kate Saenko\\
  Boston University\\
  \texttt{saenko@bu.edu}\\
  \and
  Trevor Darrell\\
  University of California, Berkeley\\
  \texttt{trevor@eecs.berkeley.edu}\\
}

\maketitle

\begin{abstract}
  Adversarial learning methods are a promising approach to training robust deep networks, and can generate complex samples across diverse domains.
  They also can improve recognition despite the presence of domain shift or dataset bias: several adversarial approaches to unsupervised domain adaptation have recently been introduced, which reduce the difference between the training and test domain distributions and thus improve generalization performance.
  Prior generative approaches show compelling visualizations, but are not optimal on discriminative tasks and can be limited to smaller shifts.
  Prior discriminative approaches could handle larger domain shifts, but imposed tied weights on the model and did not exploit a GAN-based loss.
  We first outline a novel generalized framework for adversarial adaptation, which subsumes recent state-of-the-art approaches as special cases, and we use this generalized view to better relate the prior approaches.
  We propose a previously unexplored instance of our general framework which combines discriminative modeling, untied weight sharing, and a GAN loss, which we call Adversarial Discriminative Domain Adaptation (ADDA).
  We show that ADDA is more effective yet considerably simpler than competing domain-adversarial methods, and demonstrate the promise of our approach by exceeding state-of-the-art unsupervised adaptation results on standard cross-domain digit classification tasks and a new more difficult cross-modality object classification task.
\end{abstract}

\section{Introduction}

\begin{figure}[t]
  \centering
  \includegraphics[width=\columnwidth]{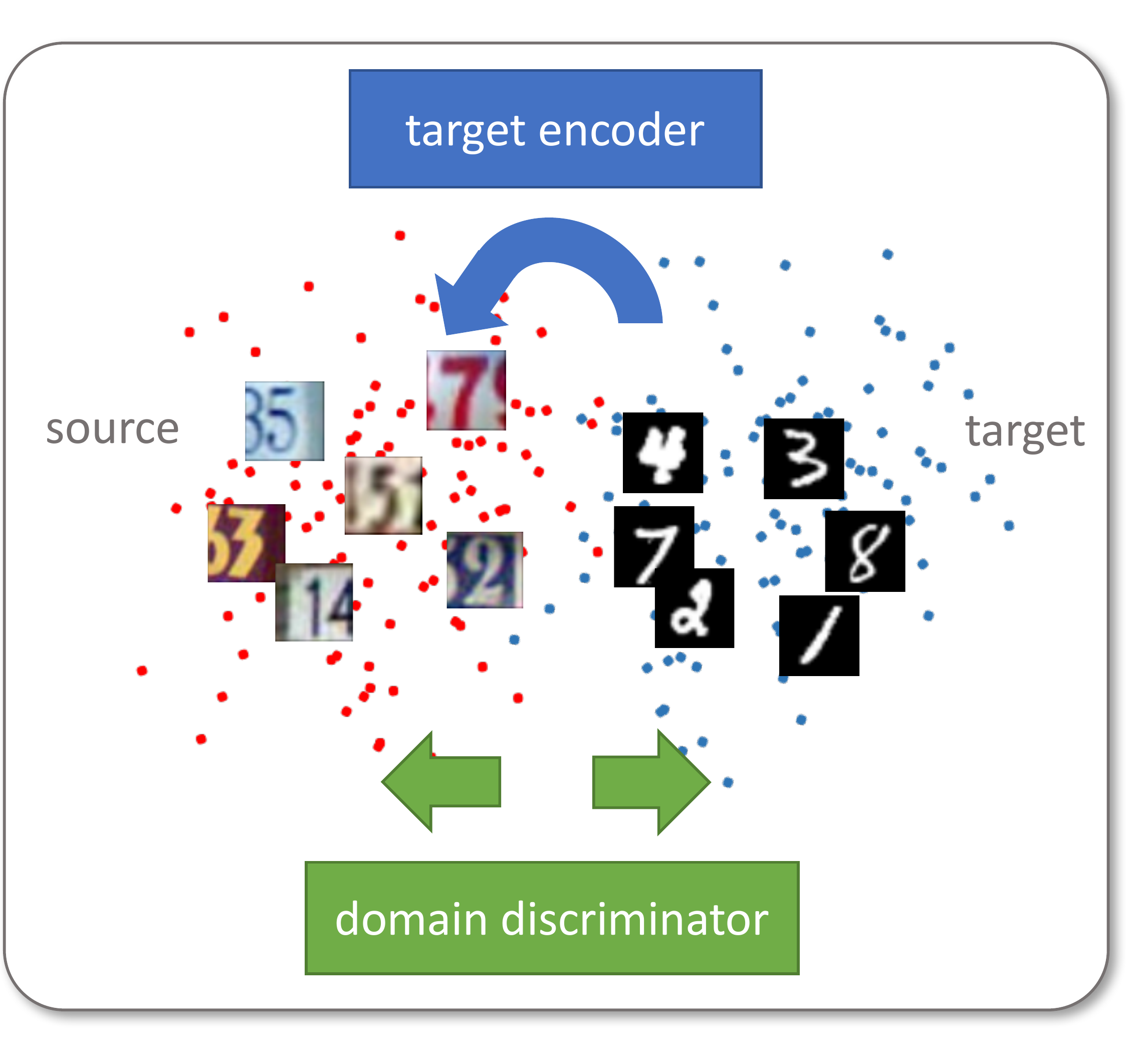}
  \caption{We propose an improved unsupervised domain adaptation method that combines adversarial learning with discriminative feature learning. Specifically, we learn a discriminative mapping of target images to the source feature space (target encoder) by fooling a domain discriminator that tries to distinguish the encoded target images from source examples.}
  \label{fig:concept}
\end{figure}

Deep convolutional networks, when trained on large-scale datasets, can learn representations which are generically usefull across a variety of tasks and visual domains~\cite{donahue2014decaf,yosinski2014transferable}.
However, due to a phenomenon known as \emph{dataset bias} or \emph{domain shift}~\cite{gretton2009}, recognition models trained along with these representations on one large dataset do not generalize well to novel datasets and tasks~\cite{torralba_cvpr11,donahue2014decaf}.
The typical solution is to further fine-tune these networks on task-specific datasets---however, it is often prohibitively difficult and expensive to obtain enough labeled data to properly fine-tune the large number of parameters employed by deep multilayer networks.

Domain adaptation methods attempt to mitigate the harmful effects of domain shift.
Recent domain adaptation methods learn deep neural transformations that map both domains into a common feature space.
This is generally achieved by optimizing the representation to minimize some measure of domain shift such as maximum mean discrepancy~\cite{tzengArxiv15, long2015learning} or correlation distances~\cite{sun_aaai16,sun_taskcv16}. An alternative is to reconstruct the target domain from the source representation~\cite{ghifary2016deep}.

Adversarial adaptation methods have become an increasingly popular incarnation of this type of approach which seeks to minimize an approximate domain discrepancy distance through an adversarial objective with respect to a domain discriminator.
These methods are closely related to generative adversarial learning~\cite{goodfellow_nips14}, which pits two networks against each other---a generator and a discriminator. The generator is trained to produce images in a way that confuses the discriminator, which in turn tries to distinguish them from real image examples. 
In domain adaptation, this principle has been employed to ensure that the network cannot distinguish between the distributions of its training and test domain examples~\cite{ganin_icml15,tzeng_iccv15,liu_arxiv16}.
However, each algorithm makes different design choices such as whether to use a generator, which loss function to employ, or whether to share weights across domains. For example,~\cite{ganin_icml15,tzeng_iccv15} share weights and learn a symmetric mapping of both source and target images to the shared feature space, while~\cite{liu_arxiv16} decouple some layers thus learning a partially asymmetric mapping.

In this work, we propose a novel unified framework for adversarial domain adaptation, allowing us to effectively examine the different factors of variation between the existing approaches and clearly view the similarities they each share.
Our framework unifies design choices such as weight-sharing, base models, and adversarial losses and subsumes previous work, while also facilitating the design of novel instantiations that improve upon existing ones.

In particular, we observe that generative modeling of input image distributions is not necessary, as the ultimate task is to learn a discriminative representation. On the other hand, asymmetric mappings can better model the difference in low level features than symmetric ones.
We therefore propose a previously unexplored unsupervised adversarial adaptation method, Adversarial Discriminative Domain Adaptation (ADDA), illustrated in Figure~\ref{fig:concept}. ADDA first learns a discriminative representation using the labels in the source domain and then a separate encoding that maps the target data to the same space using an asymmetric mapping learned through a domain-adversarial loss.
Our approach is simple yet surprisingly powerful and achieves state-of-the-art visual adaptation results on the MNIST, USPS, and SVHN digits datasets. We also test its potential to bridge the gap between even more difficult cross-modality shifts, without requiring instance constraints, by transferring object classifiers from RGB color images to depth observations.

\section{Related work}

There has been extensive prior work on domain transfer learning, see e.g.,~\cite{gretton2009}.  Recent work has focused on transferring deep neural network representations from a labeled source datasets to a target domain where labeled data is sparse or non-existent. In the case of unlabeled target domains (the focus of this paper) the main strategy has been to guide feature learning by minimizing the difference between the source and target feature distributions~\cite{ganin_icml15,tzeng_iccv15,tzengArxiv15,long2015learning,sun_taskcv16,ghifary2016deep,liu_arxiv16}.

Several methods have used the Maximum Mean Discrepancy (MMD)~\cite{gretton2009} loss for this purpose. MMD computes the norm of the difference between two domain means. The DDC method~\cite{tzengArxiv15} used MMD in addition to the regular classification loss on the source to learn a representation that is both discriminative and domain invariant. The Deep Adaptation Network (DAN)~\cite{long2015learning} applied MMD to layers embedded in a reproducing kernel Hilbert space, effectively matching higher order statistics of the two distributions. In contrast, the deep Correlation Alignment (CORAL)~\cite{sun_taskcv16} method proposed to match the mean and covariance of the two distributions.

Other methods have chosen an adversarial loss to minimize domain shift, learning a representation that is simultaneously discriminative of source labels while not being able to distinguish between domains. \cite{tzeng_iccv15} proposed adding a domain classifier (a single fully connected layer) that predicts the binary domain label of the inputs and designed a \textit{domain confusion} loss to encourage its prediction to be as close as possible to a uniform distribution over binary labels.
The gradient reversal algorithm (ReverseGrad) proposed in~\cite{ganin_icml15} also treats domain invariance as a binary classification problem, but directly maximizes the loss of the domain classifier by reversing its gradients. DRCN~\cite{ghifary2016deep} takes a similar approach but also learns to reconstruct target domain images.

In related work, adversarial learning has been explored for generative tasks.
The Generative Adversarial Network (GAN) method~\cite{goodfellow_nips14} is a generative deep model that pits two networks against one another: a generative model G that captures the data distribution and a discriminative model D that distinguishes between samples drawn from G and images drawn from the training data by predicting a binary label. The networks are trained jointly using backprop on the label prediction loss in a mini-max fashion: simultaneously update G to minimize the loss while also updating D to maximize the loss (fooling the discriminator). The advantage of GAN over other generative methods is that there is no need for complex sampling or inference during training; the downside is that it may be difficult to train.
GANs have been applied to generate natural images of objects, such as digits and faces, and have been extended in several ways. The BiGAN approach~\cite{donahue_arxiv16} extends GANs to also learn the inverse mapping from the image data back into the latent space, and shows that this can learn features useful for image classification tasks.
The conditional generative adversarial net (CGAN)~\cite{mirza2O14} is an extension of the GAN where both networks G and D receive an additional vector of information as input. This might contain, say, information about the class of the training example. The authors apply CGAN to generate a (possibly multi-modal) distribution of tag-vectors conditional on image features.

Recently the CoGAN~\cite{liu_arxiv16} approach applied GANs to the domain transfer problem by training two GANs to generate the source and target images respectively. The approach achieves a domain invariant feature space by tying the high-level layer parameters of the two GANs, and shows that the same noise input can generate a corresponding pair of images from the two distributions. Domain adaptation was performed by training a classifier on the discriminator output and applied to shifts between the MNIST and USPS digit datasets. However, this approach relies on the generators finding a mapping from the shared high-level layer feature space to full images in both domains. This can work well for say digits which can be difficult in the case of more distinct domains. In this paper, we observe that modeling the image distributions is not strictly necessary to achieve domain adaptation, as long as the latent feature space is domain invariant, and propose a discriminative approach.

\section{Generalized adversarial adaptation}

\begin{figure}
  \centering
  \includegraphics[width=\columnwidth]{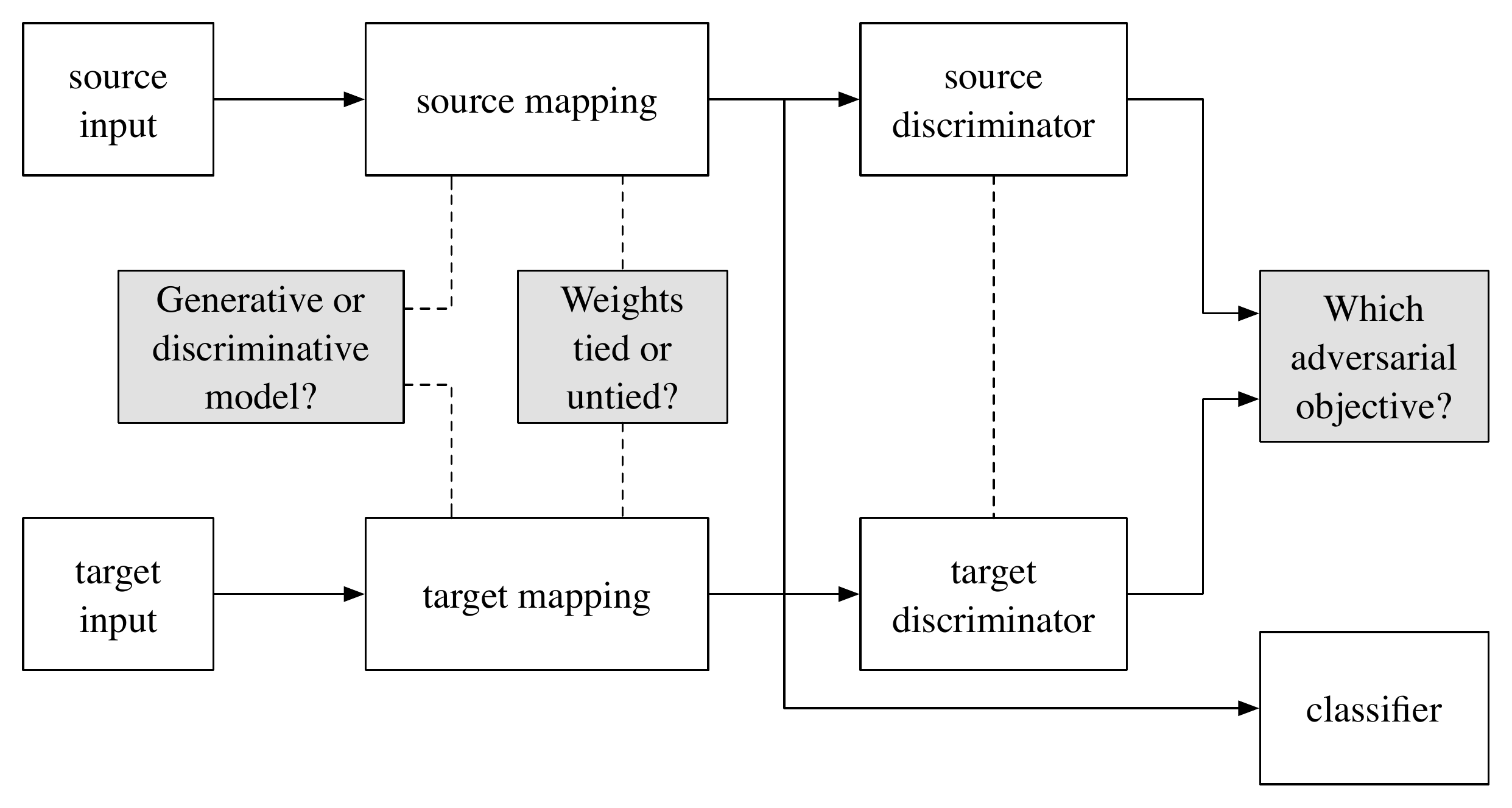}
  \caption{Our generalized architecture for adversarial domain adaptation.
  Existing adversarial adaptation methods can be viewed as instantiations of our framework with different choices regarding their properties.}
  \label{fig:overview}
\end{figure}

\begin{table*}
\centering
\begin{tabular}{lccc}
\toprule
\textbf{Method}                       & \textbf{Base model}  & \textbf{Weight sharing}  & \textbf{Adversarial loss} \\
\midrule
Gradient reversal~\cite{ganin_jmlr16} & discriminative       & shared                   & minimax                   \\
Domain confusion~\cite{tzeng_iccv15}  & discriminative       & shared                   & confusion                 \\
CoGAN~\cite{liu_arxiv16}              & generative           & unshared                 & GAN                       \\
ADDA (Ours)                           & discriminative       & unshared                 & GAN                       \\
\bottomrule
\end{tabular}
\caption{
  Overview of adversarial domain adaption methods and their various properties.
  Viewing methods under a unified framework enables us to easily propose a new adaptation method, adversarial discriminative domain adaptation (ADDA).
  }
\label{table:properties}
\end{table*}

We present a general framework for adversarial unsupervised adaptation methods. In unsupervised adaptation, we assume access to source images $\mathbf{X}_s$ and labels $Y_s$ drawn from a source domain distribution $p_s(x,y)$, as well as target images $\textbf{X}_t$ drawn from a target distribution $p_t(x,y)$, where there are no label observations.
Our goal is to learn a target representation, $M_t$ and classifier $C_t$ that can correctly classify target images into one of $K$ categories at test time, despite the lack of in domain annotations.
Since direct supervised learning on the target is not possible, domain adaptation instead learns a source representation mapping, $M_s$, along with a source classifier, $C_s$, and then learns to adapt that model for use in the target domain.

In adversarial adaptive methods, the main goal is to regularize the learning of the source and target mappings, $M_s$ and $M_t$, so as to minimize the distance between the empirical source and target mapping distributions: $M_s(X_s)$ and $M_t(X_t)$. If this is the case then the source classification model, $C_s$, can be directly applied to the target representations, elimating the need to learn a separate target classifier and instead setting, $C = C_s = C_t$.

The source classification model is then trained using the standard supervised loss below:
\begin{multline}
  \label{eq:classification}
  \min_{M_s,C} \mathcal{L}_\text{cls}(\mathbf{X}_s, Y_t) = \\
  \mathbb{E}_{(\mathbf{x}_s, y_s) \sim (\mathbf{X}_s, Y_t)} \; -\sum_{k=1}^K \mathbbm{1}_{[k = y_s]}\log C(M_s(\mathbf{x}_s))
\end{multline}

We are now able to describe our full general framework view of adversarial adaptation approaches.
We note that all approaches minimize source and target representation distances through alternating minimization between two functions. First a domain discriminator, $D$, which classifies whether a data point is drawn from the source or the target domain. Thus, $D$ is optimized according to a standard supervised loss, $\mathcal{L}_{\text{adv}_D}(\mathbf{X}_s, \mathbf{X}_t, M_s, M_t)$ where the labels indicate the origin domain, defined below:

\begin{align}
  \label{eq:adv_domain_loss}
  \begin{split}
    &\mathcal{L}_{\text{adv}_D}(\mathbf{X}_s, \mathbf{X}_t, M_s, M_t) =\\
    &\quad\quad\quad-\mathbb{E}_{\mathbf{x}_s\sim \mathbf{X}_s}[\log D(M_s(\mathbf{x}_s))] \\
    &\quad\quad\quad\quad\quad- \mathbb{E}_{\mathbf{x}_t\sim \mathbf{X}_t}[\log(1 - D(M_t(\mathbf{x}_t)))]
  \end{split}
\end{align}
Second, the source and target mappings are optimized according to a constrained adversarial objective, whose particular instantiation may vary across methods. Thus, we can derive a generic formulation for domain adversarial techniques below:
\begin{align}
  \begin{split}
    &\;\,\min_{D}   \,\;\mathcal{L}_{\text{adv}_D}(\mathbf{X}_s, \mathbf{X}_t, M_s, M_t)\\
    &\min_{M_s,M_t} \,\mathcal{L}_{\text{adv}_M}(\mathbf{X}_s, \mathbf{X}_t, D) \\
    &\,\quad\text{s.t.} \;\;\;\psi(M_s, M_t)
  \end{split}
\end{align}

In the next sections, we demonstrate the value of our framework by positioning recent domain adversarial approaches within our framework. We describe the potential mapping structure, mapping optimization constraints ($\psi(M_s, M_t)$) choices and finally choices of adversarial mapping loss, $\mathcal{L}_{\text{adv}_M}$.

\subsection{Source and target mappings}

In the case of learning a source mapping $M_s$ alone it is clear that supervised training through a latent space discriminative loss using the known labels $Y_s$ results in the best representation for final source recognition.
However, given that our target domain is unlabeled, it remains an open question how best to minimize the distance between the source and target mappings.
Thus the first choice to be made is in the particular parameterization of these mappings.

Because unsupervised domain adaptation generally considers target discriminative tasks such as classification, previous adaptation methods have generally relied on adapting discriminative models between domains~\cite{tzeng_iccv15, ganin_jmlr16}.
With a discriminative base model, input images are mapped into a feature space that is useful for a discriminative task such as image classification. For example, in the case of digit classification this may be the standard LeNet model.
However, Liu and Tuzel achieve state of the art results on unsupervised MNIST-USPS using two generative adversarial networks~\cite{liu_arxiv16}.
These generative models use random noise as input to generate samples in image space---generally, an intermediate feature of an adversarial discriminator is then used as a feature for training a task-specific classifier.

Once the mapping parameterization is determined for the source, we must decide how to parametrize the target mapping $M_t$.
In general, the target mapping almost always matches the source in terms of the specific functional layer (architecture), but different methods have proposed various regularization techniques.
All methods initialize the target mapping parameters with the source, but different methods choose different constraints between the source and target mappings, $\psi(M_s, M_t)$.
The goal is to make sure that the target mapping is set so as to minimize the distance between the source and target domains under their respective mappings, while crucially also maintaining a target mapping that is category discriminative.

Consider a layered representations where each layer parameters are denoted as, $M^{\ell}_s$ or $M^{\ell}_t$, for a given set of equivalent layers, $\{\ell_1, \dots, \ell_n\}$. Then the space of constraints explored in the literature can be described through layerwise equality constraints as follows:
\begin{equation}
	\psi(M_s, M_t) \triangleq \{\psi_{\ell_i}(M_s^{\ell_i}, M_t^{\ell_i})\}_{i \in \{1\dots n\}}
\end{equation}
where each individual layer can be constrained independently. A very common form of constraint is source and target layerwise equality:
\begin{align}
	\psi_{\ell_i}(M_s^{\ell_i}, M_t^{\ell_i}) = (M_s^{\ell_i} = M_t^{\ell_i} ).
\end{align}
It is also common to leave layers unconstrained.
These equality constraints can easily be imposed within a convolutional network framework through weight sharing.

For many prior adversarial adaptation methods~\cite{ganin_jmlr16,tzeng_iccv15}, all layers are constrained, thus enforcing exact source and target mapping consistency.
Learning a symmetric transformation reduces the number of parameters in the model and ensures that the mapping used for the target is discriminative at least when applied to the source domain.
However, this may make the optimization poorly conditioned, since the same network must handle images from two separate domains.

An alternative approach is instead to learn an asymmetric transformation with only a subset of the layers constrained, thus enforcing partial alignment.
Rozantsev et al.~\cite{rozantsev_arxiv16} showed that partially shared weights can lead to effective adaptation in both supervised and unsupervised settings.
As a result, some recent methods have favored untying weights (fully or partially) between the two domains, allowing models to learn parameters for each domain individually.

\subsection{Adversarial losses}

Once we have decided on a parametrization of $M_t$, we employ an adversarial loss to learn the actual mapping. 
There are various different possible choices of adversarial loss functions, each of which have their own unique use cases.
All adversarial losses train the adversarial discriminator using a standard classification loss, $\mathcal{L}_{\text{adv}_D}$, previously stated in Equation~\ref{eq:adv_domain_loss}.
However, they differ in the loss used to train the mapping, $\mathcal{L}_{\text{adv}_M}$.

The gradient reversal layer of~\cite{ganin_jmlr16} optimizes the mapping to maximize the discriminator loss directly:
\begin{equation}
  \mathcal{L}_{\text{adv}_M} = -\mathcal{L}_{\text{adv}_D}.
\end{equation}
This optimization corresponds to the true minimax objective for generative adversarial networks.
However, this objective can be problematic, since early on during training the discriminator converges quickly, causing the gradient to vanish.

When training GANs, rather than directly using the minimax loss, it is typical to train the generator with the standard loss function with inverted labels~\cite{goodfellow_nips14}.
This splits the optimization into two independent objectives, one for the generator and one for the discriminator, where $\mathcal{L}_{\text{adv}_D}$ remains unchanged, but $\mathcal{L}_{\text{adv}_M}$ becomes:
\begin{equation}
  \mathcal{L}_{\text{adv}_M}(\mathbf{X}_s, \mathbf{X}_t, D) = -\mathbb{E}_{\mathbf{x}_t\sim \mathbf{X}_t}[\log D(M_t(\mathbf{x}_t))].
\end{equation}
This objective has the same fixed-point properties as the minimax loss but provides stronger gradients to the target mapping.
We refer to this modified loss function as the ``GAN loss function'' for the remainder of this paper.

Note that, in this setting, we use independent mappings for source and target and learn only $M_t$ adversarially.
This mimics the GAN setting, where the real image distribution remains fixed, and the generating distribution is learned to match it.

The GAN loss function is the standard choice in the setting where the generator is attempting to mimic another unchanging distribution.
However, in the setting where both distributions are changing, this objective will lead to oscillation---when the mapping converges to its optimum, the discriminator can simply flip the sign of its prediction in response.
Tzeng et al. instead proposed the domain confusion objective, under which the mapping is trained using a cross-entropy loss function against a uniform distribution~\cite{tzeng_iccv15}:
\begin{align}
  \begin{split}
    &\mathcal{L}_{\text{adv}_M}(\mathbf{X}_s, \mathbf{X}_t, D) =\\
    &\quad\quad-\sum_{d \in \{s,t\}} \mathbb{E}_{\mathbf{x}_d\sim \mathbf{X}_d}\left[\frac{1}{2}\log D(M_d(\mathbf{x}_d))\right. \\
    &\quad\quad\quad\quad\quad\quad\left.+ \frac{1}{2}\log(1 - D(M_d(\mathbf{x}_d)))\right].
  \end{split}
\end{align}
This loss ensures that the adversarial discriminator views the two domains identically.

\section{Adversarial discriminative domain adaptation}

\begin{figure*}[t]
  \centering
  \includegraphics[width=\textwidth]{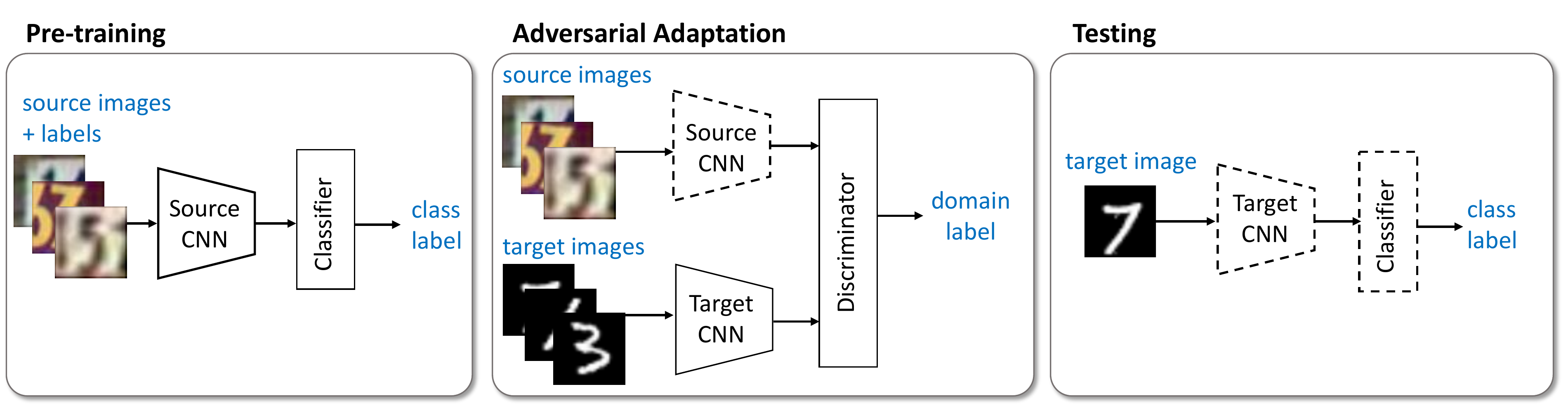}
  \caption{An overview of our proposed Adversarial Discriminative Domain Adaptation (ADDA) approach. We first pre-train a source encoder CNN using labeled source image examples. Next, we perform adversarial adaptation by learning a target encoder CNN such that a discriminator that sees encoded source and target examples cannot reliably predict their domain label. During testing, target images are mapped with the target encoder to the shared feature space and classified by the source classifier. Dashed lines indicate fixed network parameters.}
  \label{fig:method}
\end{figure*}

The benefit of our generalized framework for domain adversarial methods is that it directly enables the development of novel adaptive methods.
In fact, designing a new method has now been simplified to the space of making three design choices: whether to use a generative or discriminative base model, whether to tie or untie the weights, and which adversarial learning objective to use.
In light of this view we can summarize our method, adversarial discriminative domain adaptation (ADDA), as well as its connection to prior work, according to our choices (see Table~\ref{table:properties} ``ADDA'').
Specifically, we use a discriminative base model, unshared weights, and the standard GAN loss. We illustrate our overall sequential training procedure in Figure~\ref{fig:method}.

First, we choose a discriminative base model, as we hypothesize that much of the parameters required to generate convincing in-domain samples are irrelevant for discriminative adaptation tasks.
Most prior adversarial adaptive methods optimize directly in a discriminative space for this reason.
One counter-example is CoGANs.
However, this method has only shown dominance in settings where the source and target domain are very similar such as MNIST and USPS, and in our experiments we have had difficulty getting it to converge for larger distribution shifts.

Next, we choose to allow independent source and target mappings by untying the weights.
This is a more flexible learing paradigm as it allows more domain specific feature extraction to be learned.
However, note that the target domain has no label access, and thus without weight sharing a target model may quickly learn a degenerate solution if we do not take care with proper initialization and training procedures.
Therefore, we use the pre-trained source model as an intitialization for the target representation space and fix the source model during adversarial training.

In doing so, we are effectively learning an asymmetric mapping, in which we modify the target model so as to match the source distribution.
This is most similar to the original generative adversarial learning setting, where a generated space is updated until it is indistinguishable with a fixed real space.
Therefore, we choose the inverted label GAN loss described in the previous section.

Our proposed method, ADDA, thus corresponds to the following unconstrained optimization:

\begin{align}
  \begin{split}
    &\min_{M_s,C} \; \mathcal{L}_\text{cls}(\mathbf{X}_s, Y_s) = \\
    &\quad\quad\quad\quad-\mathbb{E}_{(\mathbf{x}_s, y_s) \sim (\mathbf{X}_s, Y_s)} \; \sum_{k=1}^K \mathbbm{1}_{[k = y_s]}\log C(M_s(\mathbf{x}_s))\\
    &\;\min_{D} \; \mathcal{L}_{\text{adv}_D}(\mathbf{X}_s, \mathbf{X}_t, M_s, M_t) =\\
    &\quad\quad\quad\quad-\mathbb{E}_{\mathbf{x}_s\sim \mathbf{X}_s}[\log D(M_s(\mathbf{x}_s))] \\
    &\quad\quad\quad\quad\quad\quad- \mathbb{E}_{\mathbf{x}_t\sim \mathbf{X}_t}[\log(1 - D(M_t(\mathbf{x}_t)))]\\
    &\min_{M_s,M_t}\mathcal{L}_{\text{adv}_M}(\mathbf{X}_s, \mathbf{X}_t, D) =\\
    &\quad\quad\quad\quad-\mathbb{E}_{\mathbf{x}_t\sim \mathbf{X}_t}[\log D(M_t(\mathbf{x}_t))].
  \end{split}
\end{align}

We choose to optimize this objective in stages.
We begin by optimizing $\mathcal{L}_\text{cls}$ over $M_s$ and $C$ by training using the labeled source data, $\mathbf{X}_s$ and $Y_s$.
Because we have opted to leave $M_s$ fixed while learning $M_t$, we can thus optimize $\mathcal{L}_{\text{adv}_D}$ and $\mathcal{L}_{\text{adv}_M}$ without revisiting the first objective term.
A summary of this entire training process is provided in Figure~\ref{fig:method}.

We note that the unified framework presented in the previous section has enabled us to compare prior domain adversarial methods and make informed decisions about the different factors of variation.
Through this framework we are able to motivate a novel domain adaptation method, ADDA, and offer insight into our design decisions.
In the next section we demonstrate promising results on unsupervised adaptation benchmark tasks, studying adaptation across digits and across modalities.

\section{Experiments}

\begin{figure*}
  \includegraphics[width=\textwidth]{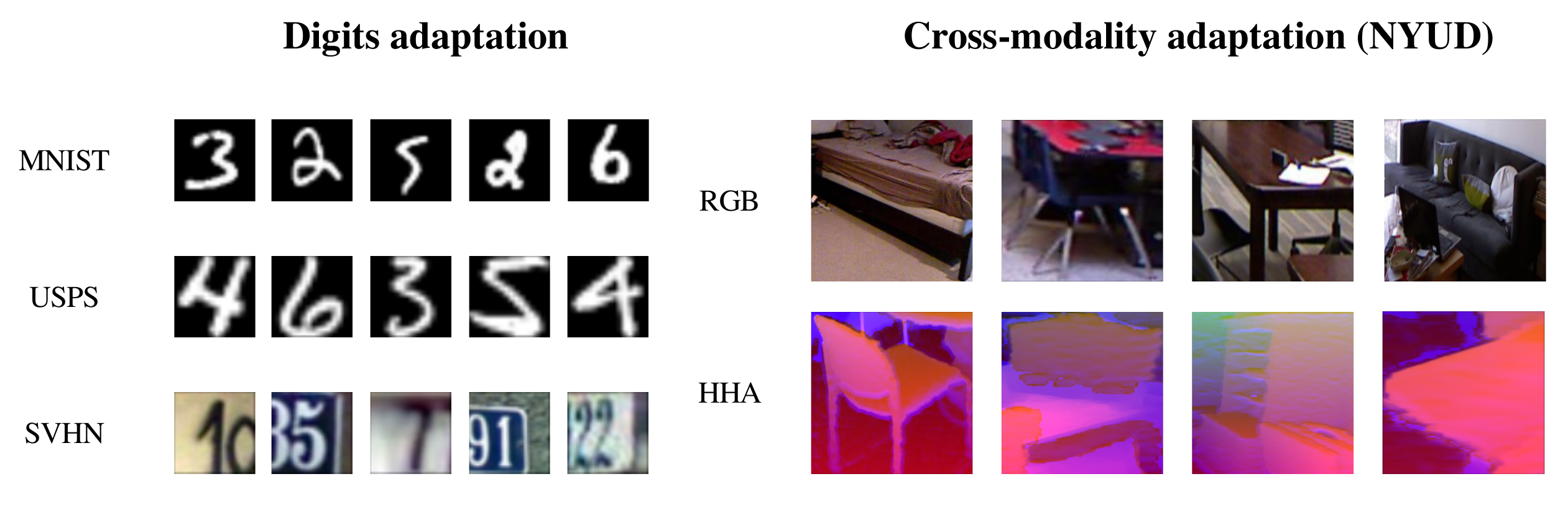}
  \caption{
    We evaluate ADDA on unsupervised adaptation across four domain shifts in two different settings.
    The first setting is adaptation between the MNIST, USPS, and SVHN datasets (left).
    The second setting is a challenging cross-modality adaptation task between RGB and depth modalities from the NYU depth dataset (right).
    }
  \label{fig:datasets}
\end{figure*}

We now evaluate ADDA for unsupervised classification adaptation across four different domain shifts.
We explore three digits datasets of varying difficulty: MNIST~\cite{lecun_ieee98}, USPS, and SVHN~\cite{netzer_nips11}.
We additionally evaluate on the NYUD~\cite{nyud2} dataset to study adaptation across modalities.
Example images from all experimental datasets are provided in Figure~\ref{fig:datasets}.

For the case of digit adaptation, we compare against multiple state-of-the-art unsupervised adaptation methods, all based upon domain adversarial learning objectives.
In 3 of 4 of our experimental setups, our method outperforms all competing approaches, and in the last domain shift studied, our approach outperforms all but one competing approach. 

We also validate our model on a real-world modality adaptation task using the NYU depth dataset.
Despite a large domain shift between the RGB and depth modalities, ADDA learns a useful depth representation without any labeled depth data and improves over the nonadaptive baseline by over 50\% (relative).

\subsection{MNIST, USPS, and SVHN digits datasets}

\begin{table*}
  \centering
  \begin{tabular}{lccc}
    \toprule
    & \textbf{MNIST} $\rightarrow$ \textbf{USPS}\hspace{0.8em} & \hspace{0.8em}\textbf{USPS} $\rightarrow$ \textbf{MNIST} & \hspace{0.4em}\textbf{SVHN} $\rightarrow$ \textbf{MNIST} \\
\textbf{Method}   & \raisebox{-.2em}{\includegraphics[height=1.1em]{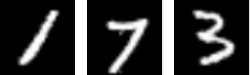}} $\rightarrow$ \raisebox{-.2em}{\includegraphics[height=1.1em]{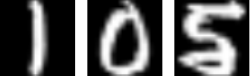}}
& \raisebox{-.2em}{\includegraphics[height=1.1em]{figs/usps}} $\rightarrow$ \raisebox{-.2em}{\includegraphics[height=1.1em]{figs/mnist}}
& \raisebox{-.2em}{\includegraphics[height=1.1em]{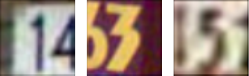}} $\rightarrow$ \raisebox{-.2em}{\includegraphics[height=1.1em]{figs/mnist}} \\
    \midrule
    Source only       & $0.752 \pm 0.016$                          & $0.571 \pm 0.017$                          & $0.601 \pm 0.011$            \\
    Gradient reversal & $0.771 \pm 0.018$                          & $0.730 \pm 0.020$                          & $0.739$~\cite{ganin_jmlr16}  \\
    Domain confusion  & $0.791 \pm 0.005$                          & $0.665 \pm 0.033$                          & $0.681 \pm 0.003$            \\
    CoGAN             & $0.912 \pm 0.008$                          & $0.891 \pm 0.008$                          & did not converge             \\
    ADDA (Ours)       & $0.894 \pm 0.002$                          & $0.901 \pm 0.008$                          & $0.760 \pm 0.018$            \\
    \bottomrule
  \end{tabular}
  \caption{Experimental results on unsupervised adaptation among MNIST, USPS, and SVHN.}
  \label{table:mnist-usps}
\end{table*}

We experimentally validate our proposed method in an unsupervised adaptation task between the MNIST~\cite{lecun_ieee98}, USPS, and SVHN~\cite{netzer_nips11} digits datasets, which consist 10 classes of digits.
Example images from each dataset are visualized in Figure~\ref{fig:datasets} and Table~\ref{table:mnist-usps}.
For adaptation between MNIST and USPS, we follow the training protocol established in~\cite{long_iccv13}, sampling 2000 images from MNIST and 1800 from USPS.
For adaptation between SVHN and MNIST, we use the full training sets for comparison against~\cite{ganin_jmlr16}.
All experiments are performed in the unsupervised settings, where labels in the target domain are withheld, and we consider adaptation in three directions: MNIST$\rightarrow$USPS, USPS$\rightarrow$MNIST, and SVHN$\rightarrow$MNIST.

For these experiments, we use the simple modified LeNet architecture provided in the Caffe source code~\cite{lecun_ieee98,jia_arxiv14}.
When training with ADDA, our adversarial discriminator consists of 3 fully connected layers: two layers with 500 hidden units followed by the final discriminator output.
Each of the 500-unit layers uses a ReLU activation function.

Results of our experiment are provided in Table~\ref{table:mnist-usps}.
On the easier MNIST and USPS shifts ADDA achieves comparable performance to the current state-of-the-art, CoGANs~\cite{liu_arxiv16}, despite being a considerably simpler model.
This provides compelling evidence that the machinery required to generate images is largely irrelevant to enabling effective adaptation.
Additionally, we show convincing results on the challenging SVHN and MNIST task in comparison to other methods, indicating that our method has the potential to generalize to a variety of settings.
In contrast, we were unable to get CoGANs to converge on SVHN and MNIST---because the domains are so disparate, we were unable to train coupled generators for them.

\subsection{Modality adaptation}

\begin{table*}
  \centering
  \footnotesize
  \setlength{\tabcolsep}{2.0pt}
  \begin{tabular}{lcccccccccccccccccccg}
    \toprule
     & \rot{bathtub} & \rot{bed} & \rot{bookshelf} & \rot{box} & \rot{chair} & \rot{counter} & \rot{desk} & \rot{door} & \rot{dresser} & \rot{garbage bin} & \rot{lamp} & \rot{monitor} & \rot{night stand} & \rot{pillow} & \rot{sink} & \rot{sofa} & \rot{table} & \rot{television} & \rot{toilet} & \rot{\textbf{overall}}\\ \midrule
    \# of instances & 19 & 96 & 87 & 210 & 611 & 103 & 122 & 129 & 25 & 55 & 144 & 37 & 51 & 276 & 47 & 129 & 210 & 33 & 17 & 2401 \\ \midrule
    Source only     & 0.000 & 0.010 & 0.011 & 0.124 & 0.188 & 0.029 & 0.041 & 0.047 & 0.000 & 0.000 & 0.069 & 0.000 & 0.039 & 0.587 & 0.000 & 0.008 & 0.010 & 0.000 & 0.000 & 0.139 \\
    ADDA (Ours)     & 0.000 & 0.146 & 0.046 & 0.229 & 0.344 & 0.447 & 0.025 & 0.023 & 0.000 & 0.018 & 0.292 & 0.081 & 0.020 & 0.297 & 0.021 & 0.116 & 0.143 & 0.091 & 0.000 & 0.211 \\ \midrule
    Train on target & 0.105 & 0.531 & 0.494 & 0.295 & 0.619 & 0.573 & 0.057 & 0.636 & 0.120 & 0.291 & 0.576 & 0.189 & 0.235 & 0.630 & 0.362 & 0.248 & 0.357 & 0.303 & 0.647 & 0.468 \\
    \bottomrule
  \end{tabular}
  \caption{Adaptation results on the NYUD~\cite{nyud2} dataset, using RGB images from the train set as source and depth images from the val set as target domains. We report here per class accuracy due to the large class imbalance in our target set (indicated in \# instances). Overall our method improves average per category accuracy from 13.9\% to 21.1\%.}
\end{table*}

\begin{figure*}
  \includegraphics[width=\textwidth]{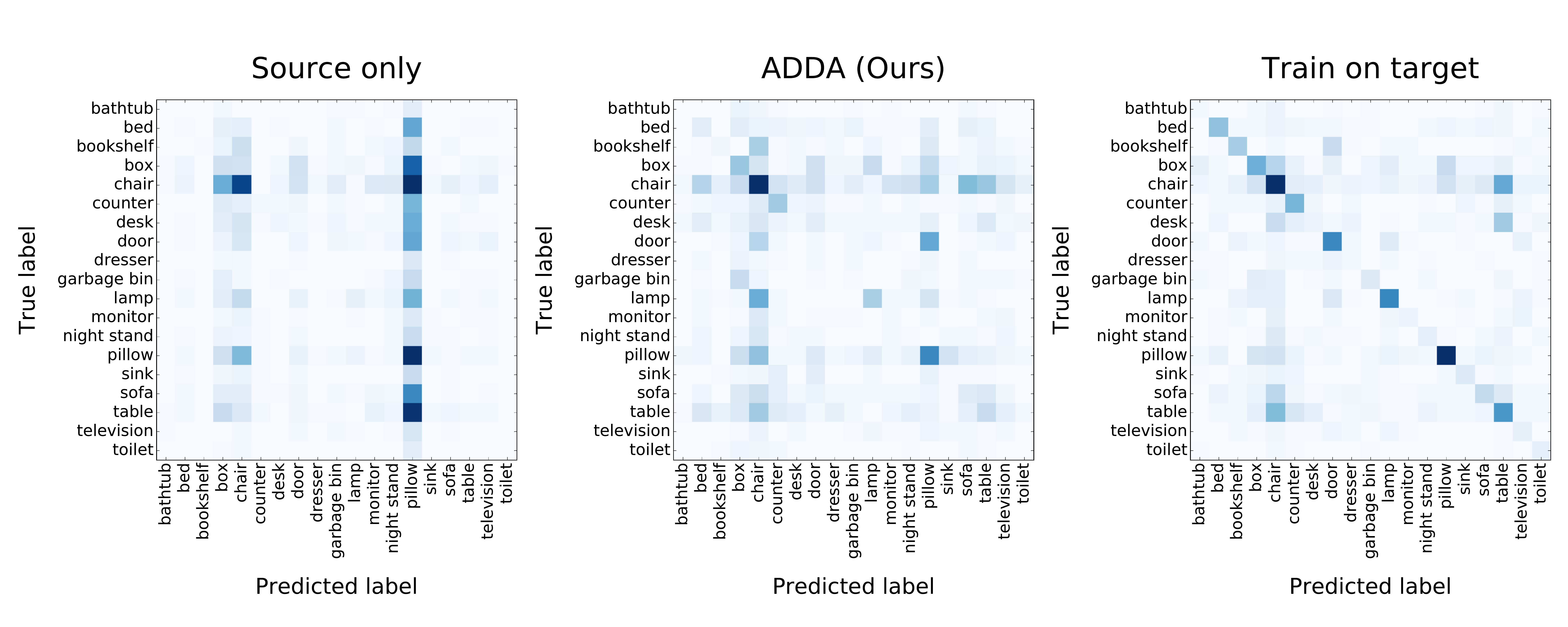}
  \caption{
    Confusion matrices for source only, ADDA, and oracle supervised target models on the NYUD RGB to depth adaptation experiment.
    We observe that our unsupervised adaptation algorithm results in a space more conducive to recognition of the most prevalent class of \texttt{chair}.
  }
  \label{fig:confusion_matrix}
\end{figure*}

We use the NYU depth dataset~\cite{nyud2}, which contains bounding box annotations for 19 object classes in 1449 images from indoor scenes. The dataset is split into a train (381 images), val (414 images) and test (654) sets. To perform our cross-modality adaptation, we
first crop out tight bounding boxes around instances of these 19 classes present in the dataset and evaluate on a 19-way classification task over object crops.
In order to ensure that the same instance is not seen in both domains, we use the RGB images from the train split as the source domain and the depth images from the val split as the target domain.
This corresponds to 2,186 labeled source images and 2,401 unlabeled target images.
Figure~\ref{fig:datasets} visualizes samples from each of the two domains.

We consider the task of adaptation between these RGB and HHA encoded depth images~\cite{gupta2014learning}, using them as source and target domains respectively.
Because the bounding boxes are tight and relatively low resolution, accurate classification is quite difficult, even when evaluating in-domain. In addition, the dataset has very few examples for certain classes, such as \texttt{toilet} and \texttt{bathtub}, which directly translates to reduced classification performance.

For this experiment, our base architecture is the VGG-16 architecture, initializing from weights pretrained on ImageNet~\cite{simonyan_arxiv14}.
This network is then fully fine-tuned on the source domain for 20,000 iterations using a batch size of 128.
When training with ADDA, the adversarial discriminator consists of three additional fully connected layers: 1024 hidden units, 2048 hidden units, then the adversarial discriminator output.
With the exception of the output, these additionally fully connected layers use a ReLU activation function.
ADDA training then proceeds for another 20,000 iterations, again with a batch size of 128.

We find that our method, ADDA, greatly improves classification accuracy for this task.
For certain categories, like \texttt{counter}, classification accuracy goes from 2.9\% under the source only baseline up to 44.7\% after adaptation.
In general, average accuracy across all classes improves significantly from 13.9\% to 21.1\%.
However, not all classes improve.
Three classes have no correctly labeled target images before adaptation, and adaptation is unable to recover performance on these classes.
Additionally, the classes of \texttt{pillow} and \texttt{nightstand} suffer performance loss after adaptation.

For additional insight on what effect ADDA has on classification, Figure~\ref{fig:confusion_matrix} plots confusion matrices before adaptation, after adaptation, and in the hypothetical best-case scenario where the target labels are present.
Examining the confusion matrix for the source only baseline reveals that the domain shift is quite large---as a result, the network is poorly conditioned and incorrectly predicts \texttt{pillow} for the majority of the dataset.
This tendency to output \texttt{pillow} also explains why the source only model achieves such abnormally high accuracy on the \texttt{pillow} class, despite poor performance on the rest of the classes.

In contrast, the classifier trained using ADDA predicts a much wider variety of classes.
This leads to decreased accuracy for the \texttt{pillow} class, but significantly higher accuracies for many of the other classes.
Additionally, comparison with the ``train on target'' model reveals that many of the mistakes the ADDA model makes are reasonable, such as confusion between the \texttt{chair} and \texttt{table} classes, indicating that the ADDA model is learning a useful representation on depth images.

\section{Conclusion}
We have proposed a unified framework for unsupervised domain adaptation techniques based on adversarial learning objectives.
Our framework provides a simplified and cohesive view by which we may understand and connect the similarities and differences between recently proposed adaptation methods.
Through this comparison, we are able to understand the benefits and key ideas from each approach and to combine these strategies into a new adaptation method, ADDA.

We present evaluation across four domain shifts for our unsupervised adaptation approach.
Our method generalizes well across a variety of tasks, achieving strong results on benchmark adaptation datasets as well as a challenging cross-modality adaptation task.
Additional analysis indicates that the representations learned via ADDA resemble features learned with supervisory data in the target domain much more closely than unadapted features, providing further evidence that ADDA is effective at partially undoing the effects of domain shift.

{
\bibliographystyle{unsrt}
\bibliography{references}

\begin{thebibliography}{10}

\bibitem{donahue2014decaf}
Jeff Donahue, Yangqing Jia, Oriol Vinyals, Judy Hoffman, Ning Zhang, Eric
  Tzeng, and Trevor Darrell.
\newblock Decaf: A deep convolutional activation feature for generic visual
  recognition.
\newblock In {\em International Conference on Machine Learning ({ICML})}, pages
  647--655, 2014.

\bibitem{yosinski2014transferable}
Jason Yosinski, Jeff Clune, Yoshua Bengio, and Hod Lipson.
\newblock How transferable are features in deep neural networks?
\newblock In {\em Neural Information Processing Systems ({NIPS})}, pages
  3320--3328, 2014.

\bibitem{gretton2009}
A.~Gretton, AJ. Smola, J.~Huang, M.~Schmittfull, KM. Borgwardt, and
  B.~Sch{\"o}lkopf.
\newblock {\em Covariate shift and local learning by distribution matching},
  pages 131--160.
\newblock MIT Press, Cambridge, MA, USA, 2009.

\bibitem{torralba_cvpr11}
Antonio Torralba and Alexei~A. Efros.
\newblock Unbiased look at dataset bias.
\newblock In {\em CVPR'11}, June 2011.

\bibitem{tzengArxiv15}
Eric Tzeng, Judy Hoffman, Ning Zhang, Kate Saenko, and Trevor Darrell.
\newblock Deep domain confusion: Maximizing for domain invariance.
\newblock {\em CoRR}, abs/1412.3474, 2014.

\bibitem{long2015learning}
Mingsheng Long and Jianmin Wang.
\newblock Learning transferable features with deep adaptation networks.
\newblock {\em International Conference on Machine Learning ({ICML})}, 2015.

\bibitem{sun_aaai16}
Baochen Sun, Jiashi Feng, and Kate Saenko.
\newblock Return of frustratingly easy domain adaptation.
\newblock In {\em Thirtieth AAAI Conference on Artificial Intelligence}, 2016.

\bibitem{sun_taskcv16}
Baochen Sun and Kate Saenko.
\newblock Deep {CORAL:} correlation alignment for deep domain adaptation.
\newblock In {\em ICCV workshop on Transferring and Adapting Source Knowledge
  in Computer Vision (TASK-CV)}, 2016.

\bibitem{ghifary2016deep}
Muhammad Ghifary, W~Bastiaan Kleijn, Mengjie Zhang, David Balduzzi, and Wen Li.
\newblock Deep reconstruction-classification networks for unsupervised domain
  adaptation.
\newblock In {\em European Conference on Computer Vision ({ECCV})}, pages
  597--613. Springer, 2016.

\bibitem{goodfellow_nips14}
Ian Goodfellow, Jean Pouget-Abadie, Mehdi Mirza, Bing Xu, David Warde-Farley,
  Sherjil Ozair, Aaron Courville, and Yoshua Bengio.
\newblock Generative adversarial nets.
\newblock In {\em Advances in Neural Information Processing Systems 27}. 2014.

\bibitem{ganin_icml15}
Yaroslav Ganin and Victor Lempitsky.
\newblock Unsupervised domain adaptation by backpropagation.
\newblock In David Blei and Francis Bach, editors, {\em Proceedings of the 32nd
  International Conference on Machine Learning (ICML-15)}, pages 1180--1189.
  JMLR Workshop and Conference Proceedings, 2015.

\bibitem{tzeng_iccv15}
Eric Tzeng, Judy Hoffman, Trevor Darrell, and Kate Saenko.
\newblock Simultaneous deep transfer across domains and tasks.
\newblock In {\em International Conference in Computer Vision (ICCV)}, 2015.

\bibitem{liu_arxiv16}
Ming{-}Yu Liu and Oncel Tuzel.
\newblock Coupled generative adversarial networks.
\newblock {\em CoRR}, abs/1606.07536, 2016.

\bibitem{donahue_arxiv16}
Jeff Donahue, Philipp Kr{\"{a}}henb{\"{u}}hl, and Trevor Darrell.
\newblock Adversarial feature learning.
\newblock {\em CoRR}, abs/1605.09782, 2016.

\bibitem{mirza2O14}
Mehdi Mirza and Simon Osindero.
\newblock Conditional generative adversarial nets.
\newblock {\em CoRR}, abs/1411.1784, 2014.

\bibitem{ganin_jmlr16}
Yaroslav Ganin, Evgeniya Ustinova, Hana Ajakan, Pascal Germain, Hugo
  Larochelle, Fran{\c{c}}ois Laviolette, Mario Marchand, and Victor Lempitsky.
\newblock Domain-adversarial training of neural networks.
\newblock {\em Journal of Machine Learning Research}, 17(59):1--35, 2016.

\bibitem{rozantsev_arxiv16}
Artem Rozantsev, Mathieu Salzmann, and Pascal Fua.
\newblock Beyond sharing weights for deep domain adaptation.
\newblock {\em CoRR}, abs/1603.06432, 2016.

\bibitem{lecun_ieee98}
Y.~LeCun, L.~Bottou, Y.~Bengio, and P.~Haffner.
\newblock Gradient-based learning applied to document recognition.
\newblock {\em Proceedings of the IEEE}, 86(11):2278--2324, November 1998.

\bibitem{netzer_nips11}
Yuval Netzer, Tao Wang, Adam Coates, Alessandro Bissacco, Bo~Wu, and Andrew~Y.
  Ng.
\newblock Reading digits in natural images with unsupervised feature learning.
\newblock In {\em NIPS Workshop on Deep Learning and Unsupervised Feature
  Learning 2011}, 2011.

\bibitem{nyud2}
Nathan Silberman, Derek Hoiem, Pushmeet Kohli, and Rob Fergus.
\newblock Indoor segmentation and support inference from rgbd images.
\newblock In {\em European Conference on Computer Vision ({ECCV})}, 2012.

\bibitem{long_iccv13}
M.~Long, J.~Wang, G.~Ding, J.~Sun, and P.~S. Yu.
\newblock Transfer feature learning with joint distribution adaptation.
\newblock In {\em 2013 IEEE International Conference on Computer Vision}, pages
  2200--2207, Dec 2013.

\bibitem{jia_arxiv14}
Yangqing Jia, Evan Shelhamer, Jeff Donahue, Sergey Karayev, Jonathan Long, Ross
  Girshick, Sergio Guadarrama, and Trevor Darrell.
\newblock Caffe: Convolutional architecture for fast feature embedding.
\newblock {\em arXiv preprint arXiv:1408.5093}, 2014.

\bibitem{gupta2014learning}
Saurabh Gupta, Ross Girshick, Pablo Arbel{\'a}ez, and Jitendra Malik.
\newblock Learning rich features from rgb-d images for object detection and
  segmentation.
\newblock In {\em European Conference on Computer Vision ({ECCV})}, pages
  345--360. Springer, 2014.

\bibitem{simonyan_arxiv14}
K.~Simonyan and A.~Zisserman.
\newblock Very deep convolutional networks for large-scale image recognition.
\newblock {\em CoRR}, abs/1409.1556, 2014.

\end{thebibliography}
}

\end{document}